\documentclass[utf8]{FrontiersinVancouver}
\usepackage{url,lineno,microtype,subcaption}
\usepackage[hidelinks]{hyperref}
\usepackage[onehalfspacing]{setspace}
\usepackage{multirow}
\usepackage{soul}
\usepackage{array}
\def\keyFont{\fontsize{8}{11}\helveticabold }
\def\firstAuthorLast{Khanmohammadi {et~al.}}
\def\Authors{R. Khanmohammadi\,$^{1}$, MM. Ghassemi\,$^{1}$, K. Verdecchia\,$^{2}$, A. I. Ghanem\,$^{3}$, L. Bing\,$^{2}$, I. J. Chetty\,$^{2}$, H. Bagher-Ebadian\,$^{2}$, F. Siddiqui\,$^{2}$, M. Elshaikh\,$^{2}$, B. Movsas\,$^{2}$ and K. Thind\,$^{2,*}$}
\def\Address{$^{1}$Department of Computer Science and Engineering, Michigan State University, USA\\
$^{2}$Department of Radiation Oncology, Henry Ford Cancer Institute, USA\\
$^{3}$Alexandria Clinical Oncology Department, Alexandria University, Egypt}
\def\corrAuthor{Kundan Thind}
\def\corrEmail{kthind1@hfhs.org}

\begin{document}
\onecolumn
\firstpage{1}

\title[Framework for assessing NLP use in clinical Radiation Oncology]{An Introduction to Natural Language Processing Techniques and Framework for Clinical Implementation in Radiation Oncology} 

\author[\firstAuthorLast ]{\Authors} 
\address{} 
\correspondance{}
\extraAuth{}
\maketitle

\begin{abstract}
Natural Language Processing (NLP) is a key technique for developing Medical Artificial Intelligence (AI) systems that leverage Electronic Health Record (EHR) data to build diagnostic and prognostic models. NLP enables the conversion of unstructured clinical text into structured data that can be fed into AI algorithms. The emergence of the transformer architecture and large language models (LLMs) has led to remarkable advances in NLP for various healthcare tasks, such as entity recognition, relation extraction, sentence similarity, text summarization, and question answering. In this article, we review the major technical innovations that underpin modern NLP models and present state-of-the-art NLP applications that employ LLMs in radiation oncology research. However, these LLMs are prone to many errors such as hallucinations, biases, and ethical violations, which necessitate rigorous evaluation and validation before clinical deployment. As such, we propose a comprehensive framework for assessing the NLP models based on their purpose and clinical fit, technical performance, bias and trust, legal and ethical implications, and quality assurance, prior to implementation in clinical radiation oncology. Our article aims to provide guidance and insights for researchers and clinicians who are interested in developing and using NLP models in clinical radiation oncology.

\tiny
 \keyFont{ \section{Keywords:} Artificial Intelligence, Radiation Oncology, Natural Language Processing, Large Language Models, Personalized Medicine} 
\end{abstract}

\section{Introduction}
Artificial intelligence (AI) is transforming healthcare by improving patient outcomes, optimizing clinical workflows, and reducing costs \citep{Bohr}. A key area where AI is being rapidly evolving is precision medicine, which tailors personalized medical care to patient's genes, environments, and lifestyles \citep{johnson2021precision}. For example, medical practitioners have the capability to detect disease-causing genetic mutations \citep{Malebary} or develop customized sedation protocols for individual patients \citep{Eghbali} by employing machine learning algorithms, which serve as a pivotal element within the realm of AI. As such, advanced AI models are being constructed within medicine for personalized disease prevention, care, and even prophylaxis. The ongoing trend of increased healthcare data collection will further aid innovative applications of these technologies in the future.

Natural Language Processing (NLP) is a subclass of AI that enables machines to understand and interpret human language. The goal of NLP is to develop algorithms and models that are capable of processing, analyzing, and generating natural language text and speech. The history of natural language processing dates back to the 1950s, when early language translation programs were developed \citep{Nwagwu}. However, limited computing power and lack of data slowed progress \citep{thompson2022computational}. The introduction of statistical methods in the 1980s led to more sophisticated language models \citep{therise}. The development of neural networks and deep learning in the 2010s led to significant advancements in NLP, culminating in the development of the famous transformer architecture in 2017 \citep{vaswani2017attention}. Riding on the capabilities of the transformer architecture, a new breed of models emerged, referred to as Large Language Models (LLMs). Notable LLMs include Bidirectional Encoder Representations from Transformers (BERT) \citep{devlin2019bert}, GPT-4 \citep{openai2023gpt4}, and ChatGPT \citep{openai2023chatgpt}. The development of these technologies has revolutionized language processing and opened new possibilities for interacting with machines using natural language.

The transformer architecture has paved the way for the development of large language models, which are now at the forefront of NLP research. Language modeling \citep{zhao2023survey}, machine translation \citep{wang2023documentlevel}, and sentiment analysis \citep{susnjak2023applying} have been shown to perform exceptionally well with these models. However, training such models is a complex and resource-intensive process requiring significant hardware and data investments. Researchers have explored the use of transfer learning to address this challenge \citep{alyafeai2020survey}, a technique that allows pre-trained models to be fine-tuned on specific tasks with limited data. Indeed, the real-world applications of NLP often begin with fine-tuning these pre-trained models to suit the specific needs of the task. Furthermore, zero-shot learning has also emerged as a promising method for generalizing new tasks without explicit training \citep{kojima2023large}. This is making these models easier to deploy, albeit evaluating their performance is becoming more challenging as input data has turned to be more expansive and complex \citep{Ahad}. As a result, researchers and practitioners in the field are facing a significant challenge as they must find ways to accurately assess the quality of these models when applied to prospective real-world data \citep{bhatt}.

Modern NLP models have shown enormous potential in healthcare, specifically in radiation oncology \citep{Yim,Savova,bitterman2021clinical,Kehl}, where precision in radiation targeting is crucial to a successful treatment \citep{santoro2022recent}. These models can analyze vast amounts of data from Electronic Health Record (EHRs) and medical literature to provide insights into disease staging, treatment options and patient outcomes \citep{netherton2021emergence}. In addition, relevant clinical information can be extracted from unstructured data sources, such as physician notes and radiology reports \citep{Wahid}, so clinicians can make better decisions about patient care. Despite this, the lack of clinical evaluation and proper validation for many of these models poses a significant challenge to their widespread adoption. The clinical evaluation of these models involves rigorous testing based on well-defined metrics and benchmark datasets. This includes assessing the potential risks and benefits in prospective use of these models, and thorough evaluation of impact on patient outcomes. The application of NLP models to the field of clinical radiation oncology has the potential to enhance its outcomes and efficiency, contingent upon the models’ ability to exhibit their validity, safety, fairness and reliability  as quantified by a rigorous evaluation framework. This requires collaboration between clinicians, data scientists, and regulatory bodies to develop detailed and robust evaluation frameworks that can ensure that these models are integrated into clinical workflows safely and effectively \citep{huynh2020artificial,PARKINSON2021S63}.

In the sections below, we review the transformer architecture, that has been foundational to development of LLMs. We will also discuss the training and fine-tuning for LLMs, and present recent application in radiation oncology. Lastly, we will identify the existing challenges and limitations for the clinical adoption of LLMs, and propose a checklist that serves as a preliminary guideline to facilitate the safe, fair, and effective application of these algorithms in clinical settings.

We employed a two-pronged approach to literature retrieval to ensure a comprehensive review of modern NLP techniques and their applicability in radiation oncology. First, we utilized ArXiv to capture the most recent advancements in the NLP field, recognizing that cutting-edge research is often disseminated through this preprint server before formal peer review. Search terms used on ArXiv included combinations of "Natural Language Processing," "Transformer Architecture," "Large Language Models," and "Clinical Applications." Subsequently, we explored PubMed to obtain NLP research applications in Radiation Oncology. Our search terms for PubMed encompassed "Natural Language Processing," "Radiation Oncology," "Electronic Health Records," and "Clinical AI Applications." While this article does not strictly adhere to the criteria of a systematic review, combining these resources ensured a timely coverage of both technical advancements and research applications, that aided in design of clinical implementation framework of these models in radiation oncology.

\section{Materials and methods}
\subsection{Literature review}
We employed a two-pronged approach to literature review to ensure a comprehensive review of modern NLP techniques and their applicability in radiation oncology. First, we utilized ArXiv to capture the most recent advancements in the NLP field, recognizing that cutting-edge research is often disseminated through this preprint server before formal peer review. Search terms used on ArXiv included combinations of "Natural Language Processing," "Transformer Architecture," "Large Language Models," and "Clinical Applications." Subsequently, we explored PubMed to obtain NLP research applications in Radiation Oncology. Our search terms for PubMed encompassed "Natural Language Processing," "Radiation Oncology," "Electronic Health Records," and "Clinical AI Applications." This literature was critical for summarizing the foundational elements of modern NLP pipelines, and their use in radiation oncology research, as well as for considerations for the design of  implementation framework of these models in clinical radiation oncology. Below, we summarize the key foundational elements of the modern NLP models. 

\subsection{Recurrent Neural Networks}
Traditionally, NLP architecture was based on Recurrent Neural Networks (RNNs). As a class of neural networks that excels at processing sequential data, RNNs provided a stepping stone for linguistic analysis, enabling models to capture temporal dynamics and dependencies within text. However, these architectures faced considerable difficulties when dealing with long-term dependencies due to issues known as vanishing or exploding gradients \cite{Mardikoraem}. To address these limitations, Hochreiter and Schmidhuber introduced in 1997 a new variant of RNNs named Long Short-Term Memory (LSTM) networks \cite{Hochreiter}. LSTMs differ from traditional RNNs primarily in their internal mechanisms, using advanced gating, cell states, and internal feedback loops.

LSTMs performed significantly more effectively at retaining and learning long-term dependencies in data. They formed the backbone of many successful models tackling a variety of NLP tasks, including machine translation, sentiment analysis, and language modeling. Yet, even with the advancements brought forth by LSTMs, they weren't exempt from challenges. Their recurrent nature implied sequential computation, limiting parallelization and computational efficiency. Moreover, while LSTMs did address the issue of capturing dependencies over longer stretches of text, they weren't always perfect. This paved the way for the transformer architecture, which not only enabled parallel computation but also harnessed self-attention mechanisms to capture long-range dependencies more effectively \citep{vaswani2017attention}.

\subsection{Transformer architecture}

The transformer architecture was introduced in the 2017 paper "Attention is All You Need" by \citet{vaswani2017attention}, revolutionizing NLP by replacing RNNs with self-attention mechanisms, enabling parallel computation and capturing long-range dependencies more effectively. Fundamentally, the transformer is composed of an encoder and a decoder. The encoder processes the input sequence, creating a context-rich representation, which the decoder then uses to produce the desired output. For instance, in sequence-to-sequence tasks like translations, the encoder might process an English sentence, and the decoder would generate its French counterpart \cite{Mirshafiee}.

The transformer architecture represents a significant improvement over previous NLP algorithms, such as LSTM \cite{Hochreiter}, in two critical aspects. Firstly, transformers excel at learning context from the writing structure and capturing long range dependencies via the mechanism of multi-headed attention \citep{hernandez2021attention}. This allows for native context understanding that was previously infeasible with LSTM. Secondly, transformers facilitate the parallelization of computations during training, which results in faster and more efficient pre-training \citep{zhuang2017parallel}.

The transformer processes an input text by breaking it down into discrete elements known as "tokens." Tokens are the building blocks of text, and depending on the granularity, a token can represent an entire word (e.g., 'Hello'), a sub-word, or even a character. Unlike sequential processing, transformers deal with these tokens concurrently, which enables capturing long-range dependencies among various parts of a sentence, a critical aspect for comprehending its overall meaning. Before being processed, each token undergoes 'input embedding', converting it into a dense vector representation that captures its semantic essence. Given the parallel nature of the transformer, it lacks a native understanding of the order of tokens. This limitation is addressed by adding 'positional encoding' to the embeddings, ensuring the model discerns the sequence order.

The architecture consists of multiple layers, each comprising two primary components: a self-attention mechanism and a feed-forward network. Rather than having multiple sub-layers, each layer functions as follows: the self-attention mechanism first computes an 'attention score' between each pair of tokens in the input sequence. These scores are then used to proportionally weigh the respective values of the tokens. The sum of these weighted values forms the output of the self-attention mechanism for that layer. Following this, a feed-forward neural network, a simple two-layer perceptron, refines the output from the self-attention mechanism. Each of these two components, the self-attention mechanism and the feed-forward network, is augmented by a residual connection and layer normalization, ensuring robust gradient flow during training. Iterating this structure, the transformer executes multiple rounds of self-attention and feed-forward processing, effectively deriving context from text, cementing its position as a dominant tool for NLP tasks.

\begin{figure*}[!t]
\centering
\includegraphics[width=0.6\linewidth,keepaspectratio]{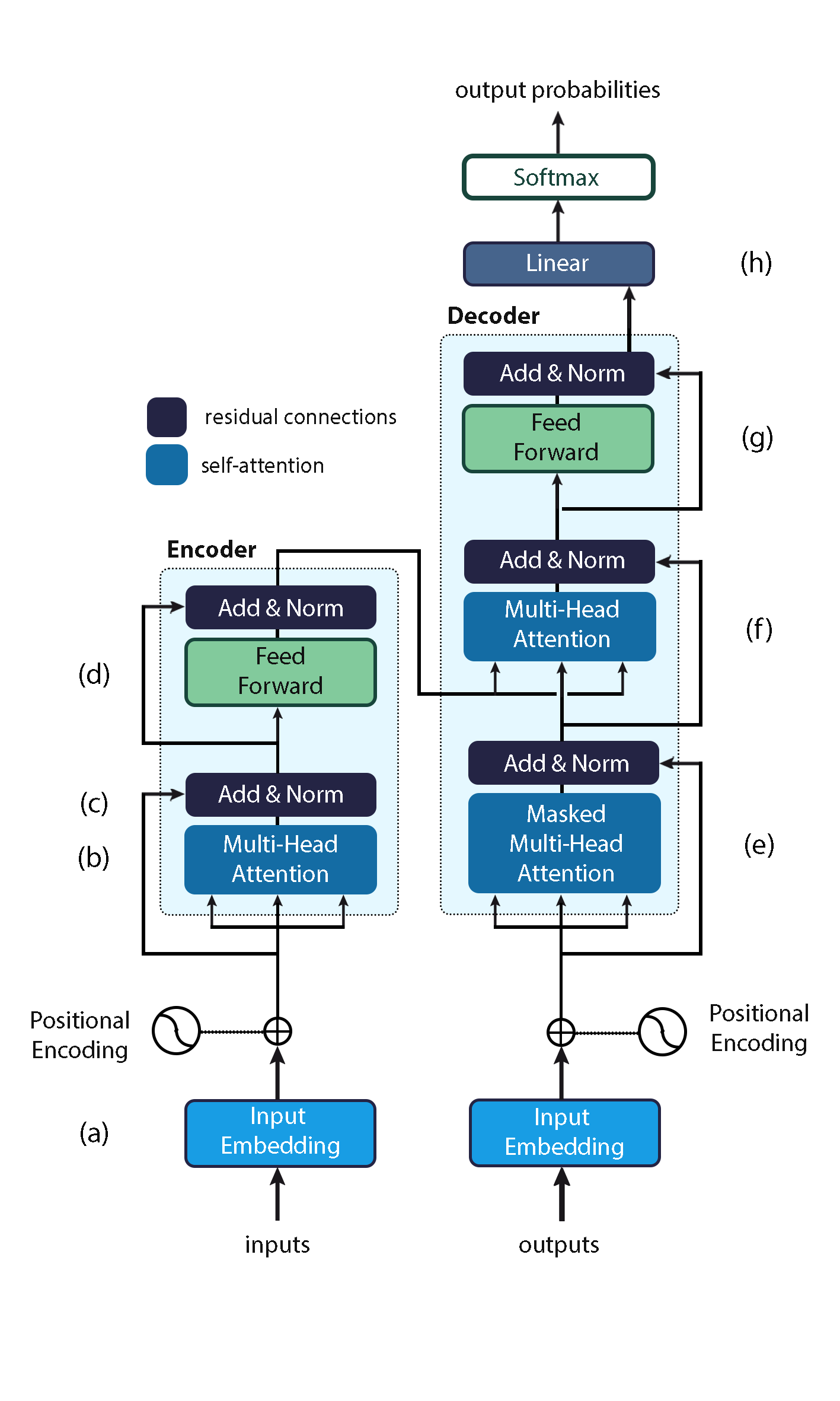}
  \caption{\citet{vaswani2017attention} illustration of the transformer architecture. First, input words are embedded and encoded based on their position in the input sequence (a). In the encoder, the Multi-head Attention enables each word to attend to other words in the sequence, capturing relationships and dependencies (b). Then, the original input embeddings are added to the self-attention outputs and normalized to integrate attended information while preserving the input (c). Next, a neural network captures complex patterns and interactions (d) before representations are added and normalized. On the other hand, the decoder's masked self-attention allows the decoder to capture dependencies among its own outputs (e). The decoder's second self-attention (f) attends to the encoded input sequence, allowing it to access the information from the input and align it with the current decoding position. Similar to the encoder, outputs are further processed with another feed-forward neural network (g), added, and normalized. Finally, the decoder applies a linear transformation (h) followed by a softmax activation function to generate the probability distribution over the vocabulary to select the next token.}
\label{fig:attention}
\end{figure*}

The transformer employs a scaled dot-product attention mechanism for self-attention. It calculates attention scores of size L x L from an input sequence of length L and dimensionality D. Each score (i, j) delineates the similarity between the i-th and j-th tokens. A softmax function then normalizes these attention scores, creating a probability distribution. The self-attention sublayer’s final output is achieved by computing a weighted sum of the input sequence, using the normalized scores as weights. This mechanism lets the model emphasize different sequence segments, thus capturing long-range dependencies, an essential trait for modeling sequences. Stacking these sublayers enables the transformer to discern language nuances and context, making it a pivotal foundation for NLP algorithms.

In summary, the profound improvement in efficiency and contextual comprehension has catalyzed the evolution of LLMs, positioning the transformer as the cornerstone of modern NLP algorithms \cite{zhao2023survey}. The ensuing section delves into LLMs' intricacies, surveys their diverse applications, and evaluates their potential in healthcare, targeting improved clinical outcomes. Figure \ref{fig:attention} visually encapsulates the discussed Transformer architecture.

\subsection{Large Language Models}
As discussed, the transformers architecture is a foundational element in the field of NLP. Its significant potential is most clearly realized in the advent and evolution of LLMs, which scale up the capabilities of transformers to unprecedented dimensions. LLMs have emerged as an evolution of Language Models (LMs), which, over the past year, have become a prominent sub-field in the domain of AI. By comprehending the patterns and structures of languages, LLMs analyze the preceding words in a sequence in order to predict the probability of a specific word sequence occurring within a given context, to predict what words appear next \citep{jing2019survey}. These were initially based on simple statistical models and have evolved into more complex algorithms based on neural networks over time. The advent of LMs coincided with the rise of machine learning, simplifying the development of powerful models \citep{Nguyen}. Several applications of this ability exist, including speech recognition, machine translation, and chatbots.  LLMs go byoned the LMs capability to predict the likelihood of a sequence of words, by generating full passages of text, offering improved performance and broader applicability. LLMs undergo extensive training on massive corpora of data, consisting of billions of parameters, and are subsequently adapted for specific downstream tasks. The potential of LLMs was first demonstrated with the release of OpenAI's Generative Pre-trained Transformer (GPT) models in 2018, followed by GPT-2 in 2019, GPT-3 in 2020, and GPT-4 in 2023. The first three iterations have 117 million, 1.5 billion, and 175 billion parameters, respectively, while no official size information exists for the fourth generation. 

LMs are categorized into three key architectures \citep{fu2023decoderonly,yang2023harnessing}. Encoder-only models like BERT process the entire input text at once, producing a context-aware vector for each token. They excel in tasks that require understanding context \citep{9443151}, as they are trained on tasks like masked language prediction. The encoder-only models are primarily designed for tasks where the relationship between parts of a text needs to be comprehensively understood, without the need for sequence generation. On the contrary, decoder-only models like GPT predict the next word based on preceding words, making them suitable for text generation tasks. This architecture is inherently designed for generating new sequences, given a context. Finally, encoder-decoder models, such as T5 \citep{raffel2020exploring} and Bidirectional Auto-Regressive Transformers (BART) \citep{lewis2019bart}, combine both approaches. They use an encoder to process the input text and a decoder to generate the output, providing a balance between context understanding and sequence generation. T5 casts all tasks as a text-to-text problem, predicting output text from input, while BART is trained to reconstruct original text from a corrupted version, allowing it to generate more coherent text. Despite this broad categorization, many models combine specific elements of the encoder and decoder architecture to fulfill specific goals. The rationale behind the adoption of a particular architecture, be it encoder-only, decoder-only, or a combination of both, revolves around the model's goal and the nature of the linguistic task it aims to excel in.

In spite of the remarkable language understanding capabilities, these models still require task-specific fine-tuning to attain optimal performance \citep{Gupta_2021}. This fine-tuning process entails exposing the model to data pertaining to the target task and modifying its parameters accordingly. The employment of these models has considerably diminished the necessity for copious amounts of task-specific data and enabled the development of highly precise and efficient NLP systems for various applications, such as language translation, question answering, and sentiment analysis \citep{9298575}. LLMs, especially those like GPT-3 and beyond, have demonstrated potential for "zero-shot" or "few-shot" learning. In a zero-shot scenario, these models attempt tasks they've never explicitly trained on, leveraging their extensive pre-training on vast datasets. For instance, ChatGPT, when used in a generalized context like chat-based applications, can generate responses in a manner that seems it was fine-tuned for a plethora of tasks. This vast and implicit understanding enables it to generalize impressively across numerous tasks. However, while this broad applicability is a transformative advancement in NLP, there might still be niche tasks where dedicated, task-specific fine-tuning delivers superior results.

Furthermore, the rise of LLMs has been pivotal in the proliferation of novel NLP applications, such as sophisticated language generation and chatbots that have the potential to redefine human-machine interaction. Their accessibility has been further augmented by the development of platforms and libraries such as Hugging Face's transformers library \citep{wolf2020huggingfaces}, which offers pre-trained models and user-friendly APIs, facilitating easier model fine-tuning and deployment. Comprehensive resources for crafting and training these models are also made available through platforms like Google's TensorFlow \citep{tensorflow2015} and Meta's PyTorch.

In the medical field, LLMs like Google's Med-PaLM \cite{Singhal2023} are making significant strides. Med-PaLM, designed for the medical domain, was the first AI to surpass the pass mark on USMLE style questions. Its successor, Med-PaLM 2 \cite{singhal2023expertlevel}, further improved accuracy in medical exams and consumer health queries. Additionally, the MED-PALM M \cite{tu2023generalist} model integrates text and imaging data, opening new frontiers in medical applications like radiomics for tumor characterization and therapy response prediction \citep{medpalm2023}.

\section{Results - research applications}
\subsection{Recent LLM Applications in Radiation Oncology}
Natural language processing has been explored for applications in cancer \cite{Kehl}, and recent advances promise improved cancer care through the use of big data such as real-world data from the EHR and the oncology information system (OIS) \cite{Honghan}. In comprehensive reviews by \citet{Yim} and \citet{bitterman2021clinical}, a diverse range of applications of NLP in the field of radiation oncology were discussed. These encompassed applications of specific information extraction from clinical notes  \cite{Liwei},  standardization of treatment planning structures \cite{MAYO20181057,Syed}, and  identification of standardized treatment locations \cite{Walker}. Furthermore, the application of NLP in toxicity data extraction was showcased, with a particular focus on the use of the clinical Text Analysis and Knowledge Extraction System (cTAKES) \cite{Savova,Hong}. The cTAKES\footnote{\url{https://ctakes.apache.org/}} is an open-source natural language processing system developed by the Mayo Clinic, designed to extract clinically relevant information from free text medical record data.. The review also highlighted the role of advanced NLP methods, specifically high-performance deep learning models, in recognizing specific radiation therapy entities \cite{Miller,Finan}, as well as their pivotal contribution in expanding cancer registries through the extraction of relevant information from clinical text \cite{SEER,ACS,ASTRO}. Additionally, the potential of NLP in identifying latent trends in radiation oncology literature was explored \cite{Rahman,OZTURK2020689}. 

The emergence of LLMs has enabled new possibilities for NLP pipelines in various domains of healthcare, and specifically in oncology and radiation oncology. We present a brief overview of some of the cutting-edge NLP applications that leverage state-of-the-art LLMs and elucidate their potential to transform clinical practice below:

\noindent\textbf{Enhancing Electronic Health Record (EHR Summarization):} LLMs can play a crucial role in summarizing electronic health records EHRs for radiation oncology. The Soft Prompt-Base Calibration (SPeC) pipeline introduced by \citet{chuang2023spec} addresses the challenge of increased output variance in EHR summarization. By incorporating SPeC, LLMs can provide more reliable and uniform summaries of vital medical information, aiding healthcare providers in making informed decisions and improving patient outcomes.

\noindent\textbf{De-identification of Radiation Oncology Data:} De-identification of sensitive medical data is of utmost importance in radiation oncology research. LLMs, such as ChatGPT and GPT-4, offer advanced named entity recognition (NER) capabilities, enabling the development of robust de-identification frameworks. \citet{Liu} proposed a GPT-4-enabled de-identification framework (DeID-GPT) that effectively masks private information from unstructured medical text, ensuring data privacy while preserving the original meaning and structure.

\noindent\textbf{Clinical Text Mining for Radiation Oncology:} LLMs, when applied to clinical text mining tasks in radiation oncology, can provide valuable insights. \citet{tang2023does} explored the use of ChatGPT for biological named entity recognition and relation extraction. To mitigate privacy concerns and improve performance, they introduced a framework that generates high-quality synthetic data with ChatGPT and fine-tunes local models for downstream tasks. This approach not only enhances performance but also reduces the time and effort required for data collection and labeling.

\noindent\textbf{Radiation Oncology Education and Knowledge Expansion:} LLMs, such as ChatGPT, hold promise as educational tools in radiation oncology. Evaluating the performance of ChatGPT in answering questions within the scope of radiation oncology exams, as demonstrated by \citet{gilson2023does}, highlights the potential of LLMs in medical education. These models can provide logical justifications and internal information, supporting learning and expanding knowledge in the field.

\noindent\textbf{Radiation Treatment Documentation and Discharge Summaries:} LLMs like ChatGPT can streamline the process of generating radiation treatment documentation and discharge summaries. \citet{pateljun} discuss the potential benefits of utilizing ChatGPT to input briefs and quickly generate comprehensive discharge summaries. While manual checking by a radiation oncologist is essential, this application of LLMs has the potential to improve efficiency and documentation quality in radiation oncology.

\subsection{Current shortcomings of NLP models}
The proliferation of LLMs has facilitated the development of conversational AI applications with widespread use cases. However, a significant challenge with these models, especially in the realm of healthcare and medicine, is their propensity for generating hallucinations—outputs that are illegitimate, irrelevant, or incorrect in relation to the original question \citep{dziri2022origin, ji}. While there have been improvements with newer models like GPT-4, hallucinations remain a pressing concern. In the medical context, such erroneous outputs can lead to misinformed decisions, potentially jeopardizing patient safety \citep{cascella2023evaluating}. As LLMs are increasingly envisioned for applications like monitoring, diagnosing, and treating patients, ensuring the reliability and accuracy of their outputs becomes paramount.

Another pivotal aspect of LLMs' performance is their embedded bias \citep{liang2021understanding,nadeem-etal-2021-stereoset} raising the need for collaboration between health professionals and data scientists to avoid encoding historical health disparities into the future. An evaluation of biases in existing NLP models used in psychiatry was presented by \citet{straw2020artificial} in a literature review of the use of NLP in mental health. The primary analysis of mental health terminology in GloVe \citep{pennington-etal-2014-glove} and Word2Vec \citep{mikolov2013efficient}  embeddings showed that significant levels of bias exist in the categories of religion, race, gender, nationality, and sexuality. Ultimately, they conclude that cross-disciplinary collaboration and communication are imperative to minimize the risks associated with health inequalities and provide recommendations for preventing these biases from occurring again due to AI and data-driven algorithmic decisions.

LLMs, trained on vast internet datasets, may harbor risks of producing offensive or discriminatory content, known as model toxicity \citep{welbl2021challenges}. Evaluations are vital to prevent biased outputs, stereotype reinforcement, or misinformation spread. This is typically done by analyzing model outputs and benchmark toxic datasets \citep{park2022detoxifying}. In addition to toxicity, other challenges such as legal concerns around AI liability in decision-making are emerging \citep{RODRIGUES2020100005}, requiring the creation of more reliable, safe, and legally compliant models. These legal concerns underscore the need for a robust framework of accountability and transparency to guide the use and deployment of these powerful AI tools, precluding any possibility for generation of any medical decision without ensuring proper review by health care providers.

Overall, the development and deployment of LLMs has become more accessible and easier to use on large cohorts of input data with rapid, fully implementable packages. On the other hand, these models generated comprehensive output, which is prone to issues as described above. This has resulted  in performance evaluation of LLMs being a challenging task compared to the previous class of NLP algorithms \citep{Bender}. Further, training and evaluating LLMs requires significant computational resources, which limits many low-resource environments, that might be in a real need for these advancements, from fully utilizing this powerful technology.  Moreover, evaluating LLMs' quality is often challenging, as nuances in natural languages require a variety of benchmarks. This challenge becomes particularly acute when LLMs are deployed in clinical settings.

Careful consideration must be made to all aspects of LLM development, performance, bias, and trust prior to implementation in the clinic. As such, rigorous evaluation and validation are required to ensure that NLP models are reliable and safe for use in the clinic. Rapid development and deployment of this technology implies a number of clinical applications that will be proposed for use in the field of radiation oncology. While these will aim to improve cancer care, a safe and methodical approach should taken to ensure a comprehensive evaluation of these algorithms prior to clinical deployment. As such, in the next section, we propose a checklist that can be utilized for step-wise evaluation of NLP models prior to clinical deployment in radiation oncology. 
\section{Results - clinical considerations}
\subsection{Framework for Clinical Implementation}

This section presents a comprehensive framework for the clinical implementation of NLP systems in radiation oncology. The framework consists of three main components: (1) evaluation of the purpose and clinical fit of the NLP system, (2) commissioning of the NLP system, and (3) quality assurance of the NLP system. The commissioning section includes sub-sections of technical performance, bias and trust, and legal and ethical scope. A graphical overview of the framework and a checklist of relevant questions for the clinical commissioning team are provided in Figure 2.

\begin{figure*}[!t]
  \centering
  \includegraphics[width=1\linewidth,keepaspectratio]{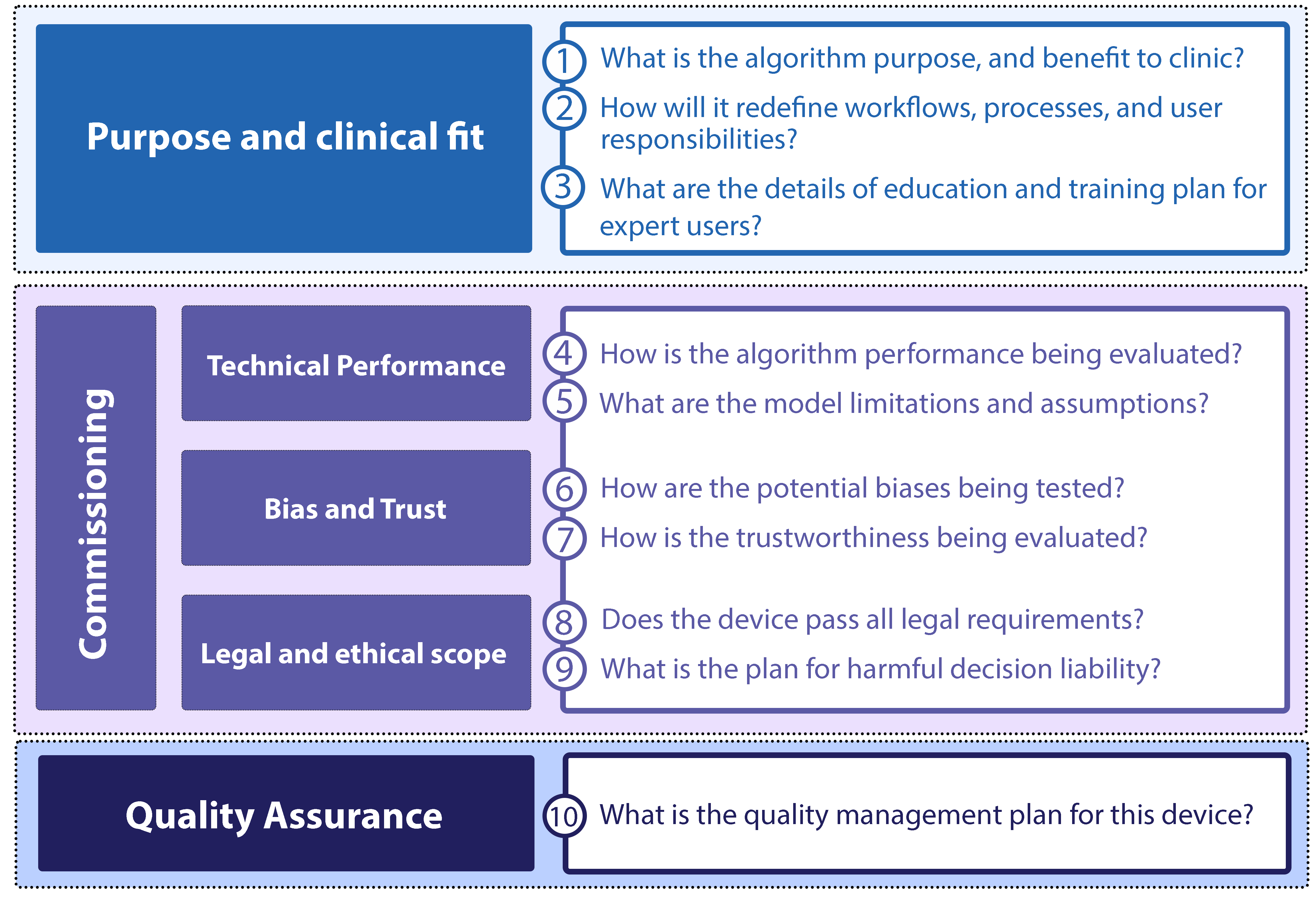}
  \caption{ Framework for assessing natural language processing algorithms before they are deployed in clinical settings. The framework consists of several categories and questions that can help evaluate the objectives, outcomes, limitations, reliability, validity, and quality control measures of the algorithms, as well as their implications for clinical practice and patient safety. }
  \label{fig:gerror}
\end{figure*}

\subsection{Purpose and clinical integration} \label{sec:app}
The first consideration for clinical implementation should be to define the purpose of the NLP model. This should include scope of the clinical problem including the specific context, appropriate NLP solution, and the expected benefit to clinic. The definition of clinical problem, and expected solution by NLP model can be general, however, it should include concrete elements that can be empirically tested and validated before proceeding to routine use. Typically, new innovations in radiation oncology are assessed for efficiency and efficacy before clinical consideration. Therefore, depending on the objective, NLP algorithm should be appraised comprehensively for efficiency and efficacy of clinical tasks and processes. A well-defined objective and a clear expectation for efficiency and efficacy are essential for thorough performance testing and consideration for clinical use.

Beyond the purpose, it is important to detail the expected change resultant from the new algorithm. As such, consideration and plan should be detailed for how the new algorithm may assist or augment the expert clinical user, and lead to a change in workflow management, and processes in the clinic. A clear education, and training plan should also be outlined for clinical users such that the expert user can use the new algorithm to its maximal capacity.  

\subsection{Commissioning} \label{sec:com}
\subsubsection{Technical performance} \label{sec:app}
NLP algorithm technical performance is evaluated on specific development datasets that is reflective of the expected clinical performance. The development dataset is usually divided into three subsets: train, development/validation, and test. The train set is used to train the initial algorithm, the validation set is used to adjust the algorithm parameters for specific tasks, and the test set is used to measure the algorithm performance. The important consideration for implementation team is to know that the testing subset should be independent from the training and tuning subsets. The level of this independence can be categorized as external validation, internal validation and cross-validation. External validation, where the testing subset comes from a different source than the training and tuning subsets, is the most rigorous evaluation method. Internal validation, where the testing subset is separated from the training and tuning subsets within a single source dataset, is the next best method. Cross-validation, where the testing subset overlaps with the training and tuning subsets, is a weaker method due to potential bias \citep{Steyerberg2016PredictionMN}. Nested Cross-Validation, however, serves as a more conservative alternative to traditional Cross-Validation, thereby reducing the risk of information leakage among different sample sub-cohorts \cite{WAINER2021115222}. In this method, Cross-Validation is performed within the training set to choose the model parameters, and an external Cross-Validation loop is used to estimate the error of the chosen model.

Once the dataset selection is validated, the algorithm performance evaluation can be appraised. Generally, performance evaluation is closely associated with the NLP task at hand. NLP tasks can be broadly categorized into classification tasks, named entity recognition, entity abstraction, summarization tasks, and question answering tasks \cite{info14040242}. Named entity recognition and entity abstraction can be evaluated using metrics of negative predictive value and positive predictive value to ascertain the sensitivity and specificity of the algorithm. The threshold for these metrics strictly depends on the use case of missing a positive case or falsely identifying a negative case as positive. Summarization tasks can be evaluated using the Rouge \citep{rouge} metrics, which evaluate the overlap between the machine-generated and human-generated summaries to ascertain the summarization's quality. Finally, question answering tasks can be evaluated using metrics like Bilingual Evaluation Understudy (BLEU) \citep{papineni}. BLEU is a widely used metric for machine translation, evaluating the quality of translated text by comparing it to a set of high-quality reference translations. It is important to understand the performance metric utilized to benchmark the algorithm, and the appropriateness to prospective clinical utilization. 

\subsubsection{Bias and trust} \label{sec:bat}
Technically, bias can be minimized by optimizing for the lowest generalization error, which is measured by the out-of-sample error and the gap between ground truth and prediction \citep{Vokinger}. This gap can be caused by model inaccuracy, sampling error, or noise. The generalization error can be reduced by choosing the right algorithm, tuning it well, and using large cohorts of diverse data. However, tuning of model should be balanced as overfitting to variance also results to poor generalization error, and poor outputs \citep{montesinos2022overfitting}. Therefore, the optimal model should capture the data's meaningful patterns without being overfitted, as demonstrated in figure \ref{fig:gerror}. Beyond the technical principle of minimizing generalization error, the algorithm should have a low bias in the domains of statistical bias, algorithmic, and societal bias \citep{NORORI2021100347,mehrabi2022survey,Olteanu}. Statistical bias can be minimized by ensuring diverse dataset with reasonable variance to ensure adequate modeling for varied outcomes. As such, there should be sufficient samples and adequate training for outcome modelling of relatively rare conditions. Algorithmic bias can be minimized by appropriate training and fine-tuning, such that underlying data patterns are appropriately represented. This step in particular ensures that all the training data is appropriately represented, and that the subsets for training, fine-tuning and testing are well defined, and have minimal leakage \citep{Naqa}. Lastly, Societal bias, which results from discrimination for, or against, a person or group, or a set of ideas or beliefs, in a way that is prejudicial or unfair. Often it shows as bias based on sub-group inequity such as gender, race, disability, sexual orientation, geography and other traits. Embedded societal bias results in poor outcomes, which can easily be amplified by the AI algorithms \citep{Panch}.  Here, it is critical that the training data is diverse, and includes  adequate representation from all groups, and sub-groups of patients. Embedded societal biases in the training data need correction, and the performance of the algorithm should be modeled and tested for different sub-groups. Even though this necessitates acquiring larger volumes of diverse data for robust analysis, which poses a significant challenge, it remains an essential task.

This is because algorithm fairness historically has not accounted for complex relationships between biological, environmental, and social factors that result in varied outcomes \citep{McCradden}. As such, this has been a widely discussed problem in AI implementation, and a few tools are publicly available for bias detection that can be utilized during algorithm testing. TCAV (Testing with Concept Activation Vectors) \citep{kim2018interpretability} is a Google-developed framework that helps to interpret and analyze the behavior of machine learning models, while audit-AI \citep{Pymetrics} and IBM's AI Fairness 360 (AIF360) \citep{Speicher,Zemel2013LearningFR} examine machine learning models to detect race, gender, location, and other sources of bias and discrimination. While these tools may be helpful overall, the testability for narrow use cases may be restrictive. If the algorithm under consideration has resulted in manuscripts and clinical trials, the CLAIM (CheckList for Artificial Intelligence in Medical Imaging) \cite{PMID33937821}, Consort-AI \cite{PMID32908283}, and CLAMP (Clinical Language Annotation, Modeling, and Processing) \cite{PMID29186491} publications can be utilized and translated to ensure adequate study organization, scientific reporting, and robust performance testing including sub-group representation \citep{Naqa,Lium3164,Soysal}. Lastly, it is important to remember that the original data used for development and testing of NLP (Natural Language Processing) pipelines may not be representative of an institution's local data. Therefore, the clinic must ensure that the algorithm ranks high on the fairness scale when applied to diverse patient populations, and as such, the tools and concepts mentioned above can aid in the clinical implementation.
\begin{figure*}[!t]
  \centering
  \includegraphics[width=0.6\linewidth,keepaspectratio]{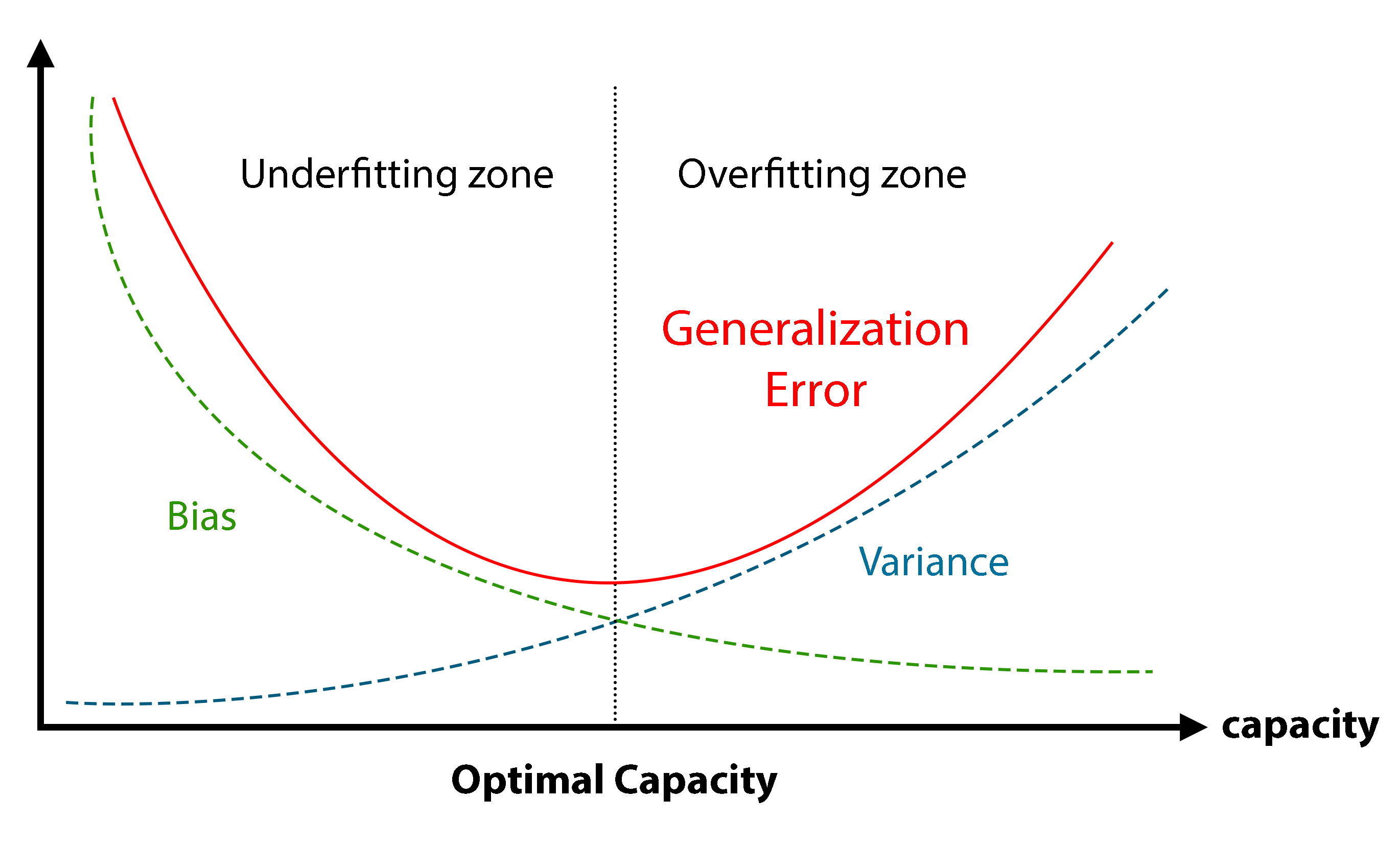}
  \caption{This figure illustrates the bias-variance tradeoff as a function of model capacity. The x-axis represents model capacity, ranging from low to high. Three key trends are displayed: Bias (descending curve), Variance (ascending curve), and Generalization Error (U-shaped curve). In the area of low capacity, the model is prone to underfitting, represented by high bias and low variance. As we move towards higher capacities, the model starts to overfit, demonstrated by low bias but high variance. The optimal model capacity is depicted where the generalization error is minimized, balancing bias and variance. This point represents the most effective model complexity for preventing both underfitting and overfitting, thus achieving optimal performance on unseen data.}
  \label{fig:gerror}
\end{figure*}
Clinician trust in NLP algorithm should be evaluated next on the path toward clinical implementation. Trust is a psychological mechanism that deals with the uncertainty between known and unknown, where the algorithm transparency, predictability, and fairness can play a large role in the trustworthiness of clinicians \citep{Asan,Varkey}. Algorithm fairness has been discussed in the previous section, where minimization of bias is critical. The transparency of the algorithm is related to the explainability and interpretability of the results, so that the algorithm can map the output to a selection of inputs \citep{Linardatos2020ExplainableAA}. In general, the more advanced an algorithm is, the lower the explainability \citep{Herm_2023}. However, to promote trust, explainability, and interpretability is increasingly being incorporated in the more advanced algorithms \citep{SHIN2021102551}. Beyond the explainability, the trust in the algorithm is largely based on the predictability of outcomes, especially when faced with conflicting inputs.

Similar to the interpretability approaches for machine learning described, NLP methods also fall into the categories of model-specific and model-agnostic interpretability \cite{carrillo2021individual}. For instance, the attention mechanism itself provides some level of model-specific interpretability. They offer a glimpse into the working of the model by illustrating the importance of different words or phrases in the input for the model's decision-making \citep{vaswani2017attention}. This way, clinicians can potentially understand which parts of a patient's history or report the algorithm deemed significant. On the model-agnostic side, techniques such as LIME \citep{c3ai_lime} (Locally Interpretable Model-agnostic Explanations) and SHAP (SHapley additive exPlanations) \citep{lundberg2017unified} have also been applied to NLP. For example, LIME can provide insight into the model's decisions by perturbing the input and observing the model's output, thereby explaining individual predictions \citep{Tulio}. SHAP values can provide a global view of feature importance across all predictions, demonstrating how much each feature (or words in a textual modality) contributes to the model's decisions. For example, a study by \citet{fetal} utilized SHAP for interpretability in clinical NLP tasks, demonstrating the most significant sound features in predicting fetal biological sex using Phonocardiogram signals. These methods enhance the transparency of NLP algorithms and provide clinicians with a better understanding of how the algorithms arrive at their conclusions. Published NLP algorithms are designed to work on specific tasks, with the embedded assumption that the training and test data are generated from a similar statistical distribution. This assumption may easily be violated in clinical scenarios, where input data may not match the statistical assumptions, and the algorithm's stability under these inputs will be directly associated with the clinician's trust in them. To test for predictability of outcomes, and an algorithm's stability, the concept of adversarial testing can be used to estimate algorithm performance under unstable inputs \citep{Goodfellow}. There are several methods for performing adversarial testing, with the evasion method perhaps being the most suitable for testing the edges of the NLP model \citep{Biggio_2013}. This testing would encompass actively modifying input data that represents the most extreme clinical scenario and analyzing the algorithm output for a) transparency - does the model explain, or interpret the outputs to specific inputs? ; b) predictability – does it show the same result every time?; c) fairness – does the output represent a sensible answer that is rooted in representation of all sub-groups within the clinical setting?  \citep{borkar2021simple}

\subsubsection{Legal and ethical considerations} \label{sec:lei}
Legal and regulatory frameworks ensure safe and ethical use of AI algorithms in healthcare. These address the potential risks and challenges, and cover three main aspects of AI development and deployment, namely: how medical devices are regulated, how health data privacy is protected, and how liability is assigned for any harm caused by faulty, erroneous, or unsafe algorithm recommendations \citep{Drabiak}. Algorithm regulation strictly follows the Food and Drug Administration (FDA) guidelines in the United States of America, and significant progress has been made by the retaliatory agency in defining standards and guiding principles for AI algorithms. NLP algorithms, being a sub-class of AI will fall under the category of Software as a Medical Device (SaMD). The FDA has established a regulatory framework IN 2019 for these devices \citep{FDAAIML}, which proposes a risk-based regulation based on the intended use of the device and the patient's risk from inaccurate output. In 2021, FDA proposed an action plan, and guiding principles in collaboration with Canada and UK to ensure safe, effective and quality SaMD use \citep{FDA2023,FDA2023b}. Recent publications have discussed the regulation of LLM based chat-bots as medical devices \cite{Gilbert2023}. Therefore, it is critical to understand the level of regulation for the device under consideration, and the regulatory implications from FDA to ensure effective clinical use. Next, health data privacy is  pivotal to ensure that patient data is kept confidential and secure. All applications and data should meet regulatory compliance under the HIPAA standards \citep{Naik}. Informed consent should also be considered to inform patients how the NLP application uses their data, which may be waived for research applications. Both the adherence to HIPAA standards, and informed consent should be evaluated for the algorithm under consideration to ensure patient data privacy, and protection.  Lastly, it is crucial to consider liability for any inaccurate decision by the NLP algorithms, whereby mechanisms for addressing any legal concerns during application use should be clarified \citep{Helen}. Despite the best intentions, flaws or errors in algorithm output can lead to patient harm, and the liability systems are incorporated into device safety to ensure patient safety and prevent unreliable, or unethical device behaviour \citep{Fotheringham,Gerke2020}. Therefore, the mechanism for addressing legal issues must be reviewed and accepted by the clinic prior to clinical implementation. 

\subsection{Quality Assurance} \label{sec:qa}
The technical and clinical performance of model is estimated during the commissioning process. A Quality Management Program (QMP) details the performance tests, frequency of testing, expected outputs, and plan of action for inadequate performance. Further it should include a quality improvement section, where the inadequate performance by the algorithm under routine use can be evaluated and discussed in detail to ensure safe and quality clinical care.  As such, QMP can be divided into routine quality assurance, and case-specific testing for quality improvement \citep{Amurao,VANDEWINCKELE202055}. Routine quality assurance must be performed periodically with a stable reference dataset every time to ensure the stability of outputs. Ideally, the reference data is a sub-set of the data that is utilized during the commissioning process, and is representative of routine clinical data and use cases. This step should also be used for clinical release of device after downtime, or minor changes, such as change in computation hardware. A comprehensive data logging system is recommended for structured collection of algorithm input, output and stability. This should a pivotal piece of the quality improvement component, whereby, the unexpected performance by the algorithm, can be traced back to the inputs, and root cause analysis can be performed to uphold safe and quality driven clinical care.  Review of case-specific performance further allows for identification of model limitations, that can facilitate future model revisions \citep{VANDEWINCKELE202055}. We believe that new and emergent use cases for the algorithm should be evaluated fully by testing purpose, clinical fit, commissioning and quality assurance plan  as outlined in section 3.1 to 3.5. This rigorous step will ensure that the algorithm performance is suitable to ensure safe and high-quality clinical care. 

\section{Conclusion}
In this article, we have discussed the recent advances and applications of NLP in radiation oncology. NLP is a powerful tool that can transform unstructured clinical narratives into structured data for medical AI systems. NLP models utilizing LLMs, and based on self-attention transformer architecture can perform multiple domain-specific tasks through transfer learning, which reduces the need for large annotated data sets and training burden. These models demonstrate remarkable performance. However, before these models can be implemented and used in routine clinical care, they need to be rigorously evaluated for their validity, functionality, viability, safety, and ethical use. We have also proposed a checklist that non-AI experts can use to assess the suitability of NLP models for their clinical needs. This checklist aptly discusses key areas of algorithm training, tuning, transparency and interpretability, bias, fairness, as well as legal and ethical concerns. Overall, these novel NLP techniques can enable the creation of more advanced AI models, which can improve patient outcomes and expedite the progress of precision medicine in radiation oncology, under appropriate ethical and technical constraints.

\bibliographystyle{Frontiers-Vancouver}
\bibliography{test}

\begin{thebibliography}{122}
\expandafter\ifx\csname natexlab\endcsname\relax\def\natexlab#1{#1}\fi
\expandafter\ifx\csname urlstyle\endcsname\relax
  \expandafter\ifx\csname doi\endcsname\relax
  \def\doi#1{doi:\discretionary{}{}{}#1}\fi \else
  \expandafter\ifx\csname doi\endcsname\relax
  \def\doi{doi:\discretionary{}{}{}\begingroup \urlstyle{rm}\Url}\fi \fi
\expandafter\ifx\csname selectlanguage\endcsname\relax
  \def\selectlanguage#1{}\fi

\bibitem[{Bohr and Memarzadeh(2020)}]{Bohr}
Bohr A, Memarzadeh K.
\newblock {\em The rise of artificial intelligence in healthcare applications\/} (2020), 25--60.
\newblock \doi{10.1016/B978-0-12-818438-7.00002-2}.

\bibitem[{Johnson et~al.(2021)Johnson, Wei, Weeraratne, Frisse, Misulis, Rhee et~al.}]{johnson2021precision}
Johnson KB, Wei WQ, Weeraratne D, Frisse ME, Misulis K, Rhee K, et~al.
\newblock Precision medicine, ai, and the future of personalized health care.
\newblock {\em Clinical and translational science\/} {\bf 14} (2021) 86--93.
\newblock \doi{10.1111/cts.12884}.

\bibitem[{Malebary and Khan(2021)}]{Malebary}
Malebary S, Khan Y.
\newblock Evaluating machine learning methodologies for identification of cancer driver genes.
\newblock {\em Scientific Reports\/} {\bf 11} (2021) 12281.
\newblock \doi{10.1038/s41598-021-91656-8}.

\bibitem[{Eghbali et~al.(2021)Eghbali, Alhanai, and Ghassemi}]{Eghbali}
Eghbali N, Alhanai T, Ghassemi M.
\newblock Patient-specific sedation management via deep reinforcement learning.
\newblock {\em Frontiers in Digital Health\/} {\bf 3} (2021).
\newblock \doi{10.3389/fdgth.2021.608893}.

\bibitem[{Nwagwu(2022)}]{Nwagwu}
[Dataset] Nwagwu W.
\newblock The rise and rise of natural language processing research, 1958-2021 (2022).
\newblock \doi{10.21203/rs.3.rs-2265814/v1}.

\bibitem[{Thompson et~al.(2022)Thompson, Greenewald, Lee, and Manso}]{thompson2022computational}
[Dataset] Thompson NC, Greenewald K, Lee K, Manso GF.
\newblock The computational limits of deep learning (2022).

\bibitem[{Hirschberg and Manning(2015)}]{therise}
Hirschberg J, Manning CD.
\newblock Advances in natural language processing.
\newblock {\em Science\/} {\bf 349} (2015) 261--266.
\newblock \doi{10.1126/science.aaa8685}.

\bibitem[{Vaswani et~al.(2017)Vaswani, Shazeer, Parmar, Uszkoreit, Jones, Gomez et~al.}]{vaswani2017attention}
[Dataset] Vaswani A, Shazeer N, Parmar N, Uszkoreit J, Jones L, Gomez AN, et~al.
\newblock Attention is all you need (2017).

\bibitem[{Devlin et~al.(2019)Devlin, Chang, Lee, and Toutanova}]{devlin2019bert}
[Dataset] Devlin J, Chang MW, Lee K, Toutanova K.
\newblock Bert: Pre-training of deep bidirectional transformers for language understanding (2019).

\bibitem[{OpenAI(2023{\natexlab{a}})}]{openai2023gpt4}
[Dataset] OpenAI.
\newblock Gpt-4 technical report (2023{\natexlab{a}}).

\bibitem[{OpenAI(2023{\natexlab{b}})}]{openai2023chatgpt}
[Dataset] OpenAI.
\newblock {ChatGPT: OpenAI's Language Model}.
\newblock \url{https://openai.com/research/chatgpt} (2023{\natexlab{b}}).

\bibitem[{Zhao et~al.(2023)Zhao, Zhou, Li, Tang, Wang, Hou et~al.}]{zhao2023survey}
[Dataset] Zhao WX, Zhou K, Li J, Tang T, Wang X, Hou Y, et~al.
\newblock A survey of large language models (2023).

\bibitem[{Wang et~al.(2023)Wang, Lyu, Ji, Zhang, Yu, Shi et~al.}]{wang2023documentlevel}
[Dataset] Wang L, Lyu C, Ji T, Zhang Z, Yu D, Shi S, et~al.
\newblock Document-level machine translation with large language models (2023).

\bibitem[{Susnjak(2023)}]{susnjak2023applying}
[Dataset] Susnjak T.
\newblock Applying bert and chatgpt for sentiment analysis of lyme disease in scientific literature (2023).

\bibitem[{Alyafeai et~al.(2020)Alyafeai, AlShaibani, and Ahmad}]{alyafeai2020survey}
[Dataset] Alyafeai Z, AlShaibani MS, Ahmad I.
\newblock A survey on transfer learning in natural language processing (2020).

\bibitem[{Kojima et~al.(2023)Kojima, Gu, Reid, Matsuo, and Iwasawa}]{kojima2023large}
[Dataset] Kojima T, Gu SS, Reid M, Matsuo Y, Iwasawa Y.
\newblock Large language models are zero-shot reasoners (2023).

\bibitem[{Ahad et~al.(2023)Ahad, TALPUR, and Jumani}]{Ahad}
Ahad A, TALPUR M, Jumani A.
\newblock Natural language processing challenges and issues: A literature review.
\newblock {\em GAZI UNIVERSITY JOURNAL OF SCIENCE\/} {\bf 36} (2023).
\newblock \doi{10.35378/gujs.1032517}.

\bibitem[{Bhatt et~al.(2021)Bhatt, Jain, Dandapat, and Sitaram}]{bhatt}
Bhatt S, Jain R, Dandapat S, Sitaram S.
\newblock A case study of efficacy and challenges in practical human-in-loop evaluation of {NLP} systems using checklist.
\newblock {\em Proceedings of the Workshop on Human Evaluation of NLP Systems (HumEval)\/} (Online: Association for Computational Linguistics) (2021), 120--130.

\bibitem[{Yim et~al.(2016)Yim, Yetisgen, Harris, and Kwan}]{Yim}
Yim WW, Yetisgen M, Harris W, Kwan S.
\newblock Natural language processing in oncology: A review.
\newblock {\em JAMA oncology\/} {\bf 2} (2016).
\newblock \doi{10.1001/jamaoncol.2016.0213}.

\bibitem[{Savova et~al.(2019)Savova, Danciu, Alamudun, Miller, Lin, Bitterman et~al.}]{Savova}
Savova G, Danciu I, Alamudun F, Miller T, Lin C, Bitterman D, et~al.
\newblock Use of natural language processing to extract clinical cancer phenotypes from electronic medical records.
\newblock {\em Cancer Research\/} {\bf 79} (2019) canres.0579.2019.
\newblock \doi{10.1158/0008-5472.CAN-19-0579}.

\bibitem[{Bitterman et~al.(2021)Bitterman, Miller, Mak, and Savova}]{bitterman2021clinical}
Bitterman DS, Miller TA, Mak RH, Savova GK.
\newblock Clinical natural language processing for radiation oncology: A review and practical primer.
\newblock {\em International journal of radiation oncology, biology, physics\/} {\bf 110} (2021) 641--655.
\newblock \doi{10.1016/j.ijrobp.2021.01.044}.

\bibitem[{Kehl et~al.(2020)Kehl, Xu, Lepisto, Elmarakeby, Hassett, Van~Allen et~al.}]{Kehl}
Kehl KL, Xu W, Lepisto E, Elmarakeby H, Hassett MJ, Van~Allen EM, et~al.
\newblock Natural language processing to ascertain cancer outcomes from medical oncologist notes.
\newblock {\em JCO Clinical Cancer Informatics\/}  (2020) 680--690.
\newblock \doi{10.1200/CCI.20.00020}.
\newblock PMID: 32755459.

\bibitem[{Santoro et~al.(2022)Santoro, Strolin, Paolani, Della~Gala, Bartoloni, Giacometti et~al.}]{santoro2022recent}
Santoro M, Strolin S, Paolani G, Della~Gala G, Bartoloni A, Giacometti C, et~al.
\newblock Recent applications of artificial intelligence in radiotherapy: Where we are and beyond.
\newblock {\em Applied Sciences\/} {\bf 12} (2022) 3223.
\newblock \doi{10.3390/app12073223}.

\bibitem[{Netherton et~al.(2021)Netherton, Cardenas, Rhee, Court, and Beadle}]{netherton2021emergence}
Netherton TJ, Cardenas CE, Rhee DJ, Court LE, Beadle BM.
\newblock The emergence of artificial intelligence within radiation oncology treatment planning.
\newblock {\em Oncology\/} {\bf 99} (2021) 124--134.
\newblock \doi{10.1159/000512172}.

\bibitem[{Wahid et~al.(2022)Wahid, Glerean, Sahlsten, Jaskari, Kaski, Naser et~al.}]{Wahid}
Wahid K, Glerean E, Sahlsten J, Jaskari J, Kaski K, Naser M, et~al.
\newblock Artificial intelligence for radiation oncology applications using public datasets.
\newblock {\em Seminars in Radiation Oncology\/} {\bf 32} (2022) 400--414.
\newblock \doi{10.1016/j.semradonc.2022.06.009}.

\bibitem[{Huynh et~al.(2020)Huynh, Hosny, Guthier, Bitterman, Petit, Haas-Kogan et~al.}]{huynh2020artificial}
Huynh E, Hosny A, Guthier C, Bitterman DS, Petit SF, Haas-Kogan DA, et~al.
\newblock Artificial intelligence in radiation oncology.
\newblock {\em Nature reviews. Clinical oncology\/} {\bf 17} (2020) 771--781.
\newblock \doi{10.1038/s41571-020-0417-8}.

\bibitem[{Parkinson et~al.(2021)Parkinson, Matthams, Foley, and Spezi}]{PARKINSON2021S63}
Parkinson C, Matthams C, Foley K, Spezi E.
\newblock Artificial intelligence in radiation oncology: A review of its current status and potential application for the radiotherapy workforce.
\newblock {\em Radiography\/} {\bf 27} (2021) S63--S68.
\newblock \doi{https://doi.org/10.1016/j.radi.2021.07.012}.
\newblock The Future Role of the Radiographer.

\bibitem[{Mardikoraem et~al.(2023)Mardikoraem, Wang, Pascual, and Woldring}]{Mardikoraem}
Mardikoraem M, Wang Z, Pascual N, Woldring D.
\newblock {Generative models for protein sequence modeling: recent advances and future directions}.
\newblock {\em Briefings in Bioinformatics\/} {\bf 24} (2023) bbad358.
\newblock \doi{10.1093/bib/bbad358}.

\bibitem[{Hochreiter and Schmidhuber(1997)}]{Hochreiter}
Hochreiter S, Schmidhuber J.
\newblock Long short-term memory.
\newblock {\em Neural Comput.\/} {\bf 9} (1997) 1735–1780.
\newblock \doi{10.1162/neco.1997.9.8.1735}.

\bibitem[{Khanmohammadi et~al.(2023)Khanmohammadi, Mirshafiee, Rezaee~Jouryabi, and Mirroshandel}]{Mirshafiee}
Khanmohammadi R, Mirshafiee MS, Rezaee~Jouryabi Y, Mirroshandel SA.
\newblock Prose2poem: The blessing of transformers in translating prose to persian poetry.
\newblock {\em ACM Trans. Asian Low-Resour. Lang. Inf. Process.\/} {\bf 22} (2023).
\newblock \doi{10.1145/3592791}.

\bibitem[{Hernández and Amigó(2021)}]{hernandez2021attention}
Hernández A, Amigó JM.
\newblock Attention mechanisms and their applications to complex systems.
\newblock {\em Entropy (Basel)\/} {\bf 23} (2021) 283.
\newblock \doi{10.3390/e23030283}.

\bibitem[{Zhuang et~al.(2017)Zhuang, Wu, Shen, Reid, and van~den Hengel}]{zhuang2017parallel}
[Dataset] Zhuang B, Wu Q, Shen C, Reid I, van~den Hengel A.
\newblock Parallel attention: A unified framework for visual object discovery through dialogs and queries (2017).

\bibitem[{Jing and Xu(2019)}]{jing2019survey}
[Dataset] Jing K, Xu J.
\newblock A survey on neural network language models (2019).

\bibitem[{Nguyen et~al.(2019)Nguyen, Dlugolinsky, Bob\'{a}k, Tran, L\'{o}pez~Garc\'{\i}a, Heredia et~al.}]{Nguyen}
Nguyen G, Dlugolinsky S, Bob\'{a}k M, Tran V, L\'{o}pez~Garc\'{\i}a A, Heredia I, et~al.
\newblock Machine learning and deep learning frameworks and libraries for large-scale data mining: A survey.
\newblock {\em Artif. Intell. Rev.\/} {\bf 52} (2019) 77–124.
\newblock \doi{10.1007/s10462-018-09679-z}.

\bibitem[{Fu et~al.(2023)Fu, Lam, Yu, So, Hu, Liu et~al.}]{fu2023decoderonly}
[Dataset] Fu Z, Lam W, Yu Q, So AMC, Hu S, Liu Z, et~al.
\newblock Decoder-only or encoder-decoder? interpreting language model as a regularized encoder-decoder (2023).

\bibitem[{Yang et~al.(2023)Yang, Jin, Tang, Han, Feng, Jiang et~al.}]{yang2023harnessing}
[Dataset] Yang J, Jin H, Tang R, Han X, Feng Q, Jiang H, et~al.
\newblock Harnessing the power of llms in practice: A survey on chatgpt and beyond (2023).

\bibitem[{Khanmohammadi et~al.(2021{\natexlab{a}})Khanmohammadi, Mirshafiee, and Allahyari}]{9443151}
Khanmohammadi R, Mirshafiee MS, Allahyari M.
\newblock Coper: a query-adaptable semantics-based search engine for persian covid-19 articles.
\newblock {\em 2021 7th International Conference on Web Research (ICWR)\/} (2021{\natexlab{a}}), 64--70.
\newblock \doi{10.1109/ICWR51868.2021.9443151}.

\bibitem[{Raffel et~al.(2020)Raffel, Shazeer, Roberts, Lee, Narang, Matena et~al.}]{raffel2020exploring}
[Dataset] Raffel C, Shazeer N, Roberts A, Lee K, Narang S, Matena M, et~al.
\newblock Exploring the limits of transfer learning with a unified text-to-text transformer (2020).

\bibitem[{Lewis et~al.(2019)Lewis, Liu, Goyal, Ghazvininejad, Mohamed, Levy et~al.}]{lewis2019bart}
[Dataset] Lewis M, Liu Y, Goyal N, Ghazvininejad M, Mohamed A, Levy O, et~al.
\newblock Bart: Denoising sequence-to-sequence pre-training for natural language generation, translation, and comprehension (2019).

\bibitem[{Gupta(2021)}]{Gupta_2021}
Gupta N.
\newblock A pre-trained vs fine-tuning methodology in transfer learning.
\newblock {\em Journal of Physics: Conference Series\/} {\bf 1947} (2021) 012028.
\newblock \doi{10.1088/1742-6596/1947/1/012028}.

\bibitem[{Ramdan et~al.(2020)Ramdan, Heryana, Arisal, Kusumo, and Pardede}]{9298575}
Ramdan A, Heryana A, Arisal A, Kusumo RBS, Pardede HF.
\newblock Transfer learning and fine-tuning for deep learning-based tea diseases detection on small datasets.
\newblock {\em 2020 International Conference on Radar, Antenna, Microwave, Electronics, and Telecommunications (ICRAMET)\/} (2020), 206--211.
\newblock \doi{10.1109/ICRAMET51080.2020.9298575}.

\bibitem[{Wolf et~al.(2020)Wolf, Debut, Sanh, Chaumond, Delangue, Moi et~al.}]{wolf2020huggingfaces}
[Dataset] Wolf T, Debut L, Sanh V, Chaumond J, Delangue C, Moi A, et~al.
\newblock Huggingface's transformers: State-of-the-art natural language processing (2020).

\bibitem[{Abadi et~al.(2015)Abadi, Agarwal, Barham, Brevdo, Chen, Citro et~al.}]{tensorflow2015}
[Dataset] Abadi M, Agarwal A, Barham P, Brevdo E, Chen Z, Citro C, et~al.
\newblock {TensorFlow}: Large-scale machine learning on heterogeneous systems (2015).
\newblock Software available from tensorflow.org.

\bibitem[{Singhal et~al.(2023{\natexlab{a}})Singhal, Azizi, Tu, Mahdavi, Wei, Chung et~al.}]{Singhal2023}
Singhal K, Azizi S, Tu T, Mahdavi SS, Wei J, Chung HW, et~al.
\newblock Large language models encode clinical knowledge.
\newblock {\em Nature\/} {\bf 620} (2023{\natexlab{a}}) 172--180.
\newblock \doi{10.1038/s41586-023-06291-2}.

\bibitem[{Singhal et~al.(2023{\natexlab{b}})Singhal, Tu, Gottweis, Sayres, Wulczyn, Hou et~al.}]{singhal2023expertlevel}
[Dataset] Singhal K, Tu T, Gottweis J, Sayres R, Wulczyn E, Hou L, et~al.
\newblock Towards expert-level medical question answering with large language models (2023{\natexlab{b}}).

\bibitem[{Tu et~al.(2023)Tu, Azizi, Driess, Schaekermann, Amin, Chang et~al.}]{tu2023generalist}
[Dataset] Tu T, Azizi S, Driess D, Schaekermann M, Amin M, Chang PC, et~al.
\newblock Towards generalist biomedical ai (2023).

\bibitem[{Wu et~al.(2022)Wu, Wang, Wu, Francis, Chang, Shavick et~al.}]{Honghan}
Wu H, Wang M, Wu J, Francis F, Chang YH, Shavick A, et~al.
\newblock A survey on clinical natural language processing in the united kingdom from 2007 to 2022.
\newblock {\em npj Digital Medicine\/} {\bf 5} (2022) 186.
\newblock \doi{10.1038/s41746-022-00730-6}.

\bibitem[{Wang et~al.(2019)Wang, Luo, Wang, Wampfler, Yang, and Liu}]{Liwei}
Wang L, Luo L, Wang Y, Wampfler J, Yang P, Liu H.
\newblock Information extraction for populating lung cancer clinical research data (2019), vol. 2019, 1--2.
\newblock \doi{10.1109/ICHI.2019.8904601}.

\bibitem[{MAY(2018)}]{MAYO20181057}
American association of physicists in medicine task group 263: Standardizing nomenclatures in radiation oncology.
\newblock {\em International Journal of Radiation Oncology*Biology*Physics\/} {\bf 100} (2018) 1057--1066.
\newblock \doi{https://doi.org/10.1016/j.ijrobp.2017.12.013}.

\bibitem[{Syed et~al.(2020)Syed, Sleeman, Ivey, Hagan, Palta, Kapoor et~al.}]{Syed}
Syed K, Sleeman W, Ivey K, Hagan M, Palta J, Kapoor R, et~al.
\newblock Integrated natural language processing and machine learning models for standardizing radiotherapy structure names.
\newblock {\em Healthcare\/} {\bf 8} (2020).
\newblock \doi{10.3390/healthcare8020120}.

\bibitem[{Walker et~al.(2019)Walker, Soysal, and Qi}]{Walker}
Walker G, Soysal E, Qi W.
\newblock Development of a natural language processing tool to extract radiation treatment sites.
\newblock {\em Cureus\/} {\bf 11} (2019).
\newblock \doi{10.7759/cureus.6010}.

\bibitem[{Hong et~al.(2020)Hong, Fairchild, Tanksley, Palta, and Tenenbaum}]{Hong}
Hong J, Fairchild A, Tanksley J, Palta M, Tenenbaum J.
\newblock Natural language processing for abstraction of cancer treatment toxicities: accuracy versus human experts.
\newblock {\em JAMIA Open\/} {\bf 3} (2020).
\newblock \doi{10.1093/jamiaopen/ooaa064}.

\bibitem[{Bitterman et~al.(2020{\natexlab{a}})Bitterman, Miller, Harris, Lin, Finan, Warner et~al.}]{Miller}
Bitterman D, Miller T, Harris D, Lin C, Finan S, Warner J, et~al.
\newblock Extracting radiotherapy treatment details using neural network-based natural language processing.
\newblock {\em International Journal of Radiation Oncology*Biology*Physics\/} {\bf 108} (2020{\natexlab{a}}) e771--e772.
\newblock \doi{10.1016/j.ijrobp.2020.07.219}.

\bibitem[{Bitterman et~al.(2020{\natexlab{b}})Bitterman, Miller, Harris, Lin, Finan, Warner et~al.}]{Finan}
Bitterman D, Miller T, Harris D, Lin C, Finan S, Warner J, et~al.
\newblock Extracting relations between radiotherapy treatment details.
\newblock {\em Proceedings of the 3rd Clinical Natural Language Processing Workshop\/} (Online: Association for Computational Linguistics) (2020{\natexlab{b}}), 194--200.
\newblock \doi{10.18653/v1/2020.clinicalnlp-1.21}.

\bibitem[{{National Cancer Institute}(Year)}]{SEER}
[Dataset] {National Cancer Institute}.
\newblock {Surveillance, Epidemiology, and End Results Program (SEER)}.
\newblock [Online] (Year).
\newblock Accessed May 31, 2023.

\bibitem[{{American College of Surgeons}(2020)}]{ACS}
[Dataset] {American College of Surgeons}.
\newblock {National Cancer Database}.
\newblock [Online] (2020).
\newblock Accessed May 31, 2023.

\bibitem[{for Radiation Oncology~(ASTRO)(2020)}]{ASTRO}
[Dataset] for Radiation Oncology~(ASTRO) AS.
\newblock New registry launched to track and improve the quality of cancer care delivered in the u.s.
\newblock [Online] (2020).
\newblock Accessed May 31, 2023.

\bibitem[{Rahman et~al.(2019)Rahman, Fell, Ventz, Arfe, Vanderbeek, Trippa et~al.}]{Rahman}
Rahman R, Fell G, Ventz S, Arfe A, Vanderbeek A, Trippa L, et~al.
\newblock Deviation from the proportional hazards assumption in randomized phase 3 clinical trials in oncology: Prevalence, associated factors and implications.
\newblock {\em Clinical Cancer Research\/} {\bf 25} (2019) clincanres.3999.2018.
\newblock \doi{10.1158/1078-0432.CCR-18-3999}.

\bibitem[{Öztürk et~al.(2020)Öztürk, Özgür, Schwaller, Laino, and Ozkirimli}]{OZTURK2020689}
Öztürk H, Özgür A, Schwaller P, Laino T, Ozkirimli E.
\newblock Exploring chemical space using natural language processing methodologies for drug discovery.
\newblock {\em Drug Discovery Today\/} {\bf 25} (2020) 689--705.
\newblock \doi{https://doi.org/10.1016/j.drudis.2020.01.020}.

\bibitem[{Chuang et~al.(2023)Chuang, Tang, Jiang, and Hu}]{chuang2023spec}
Chuang YN, Tang R, Jiang X, Hu X.
\newblock Spec: A soft prompt-based calibration on mitigating performance variability in clinical notes summarization.
\newblock {\em arXiv preprint arXiv:2303.13035\/}  (2023).

\bibitem[{Liu et~al.(2023)Liu, Yu, Zhang, Wu, Cao, Dai et~al.}]{Liu}
[Dataset] Liu Z, Yu X, Zhang L, Wu Z, Cao C, Dai H, et~al.
\newblock Deid-gpt: Zero-shot medical text de-identification by gpt-4 (2023).

\bibitem[{Tang et~al.(2023)Tang, Han, Jiang, and Hu}]{tang2023does}
[Dataset] Tang R, Han X, Jiang X, Hu X.
\newblock Does synthetic data generation of llms help clinical text mining? (2023).

\bibitem[{Gilson et~al.(2023)Gilson, Safranek, Huang, Socrates, Chi, Taylor et~al.}]{gilson2023does}
Gilson A, Safranek CW, Huang T, Socrates V, Chi L, Taylor RA, et~al.
\newblock How does chatgpt perform on the united states medical licensing examination? the implications of large language models for medical education and knowledge assessment.
\newblock {\em JMIR Medical Education\/} {\bf 9} (2023) e45312.

\bibitem[{Patel and Lam(2023)}]{pateljun}
Patel S, Lam K.
\newblock {ChatGPT}: the future of discharge summaries?
\newblock {\em The Lancet Digital Health\/} {\bf 5} (2023) e107--e108.
\newblock \doi{10.1016/S2589-7500(23)00021-3}.
\newblock Publisher: Elsevier Ltd.

\bibitem[{Dziri et~al.(2022)Dziri, Milton, Yu, Zaiane, and Reddy}]{dziri2022origin}
[Dataset] Dziri N, Milton S, Yu M, Zaiane O, Reddy S.
\newblock On the origin of hallucinations in conversational models: Is it the datasets or the models? (2022).

\bibitem[{Ji et~al.(2023)Ji, Lee, Frieske, Yu, Su, Xu et~al.}]{ji}
Ji Z, Lee N, Frieske R, Yu T, Su D, Xu Y, et~al.
\newblock Survey of hallucination in natural language generation.
\newblock {\em ACM Comput. Surv.\/} {\bf 55} (2023).
\newblock \doi{10.1145/3571730}.

\bibitem[{Cascella et~al.(2023)Cascella, Montomoli, Bellini, and Bignami}]{cascella2023evaluating}
Cascella M, Montomoli J, Bellini V, Bignami E.
\newblock Evaluating the feasibility of chatgpt in healthcare: An analysis of multiple clinical and research scenarios.
\newblock {\em Journal of Medical Systems\/} {\bf 47} (2023) 1--5.

\bibitem[{Liang et~al.(2021)Liang, Wu, Morency, and Salakhutdinov}]{liang2021understanding}
[Dataset] Liang PP, Wu C, Morency LP, Salakhutdinov R.
\newblock Towards understanding and mitigating social biases in language models (2021).

\bibitem[{Nadeem et~al.(2021)Nadeem, Bethke, and Reddy}]{nadeem-etal-2021-stereoset}
Nadeem M, Bethke A, Reddy S.
\newblock {S}tereo{S}et: Measuring stereotypical bias in pretrained language models.
\newblock {\em Proceedings of the 59th Annual Meeting of the Association for Computational Linguistics and the 11th International Joint Conference on Natural Language Processing (Volume 1: Long Papers)\/} (Online: Association for Computational Linguistics) (2021), 5356--5371.
\newblock \doi{10.18653/v1/2021.acl-long.416}.

\bibitem[{Straw and Callison-Burch(2020)}]{straw2020artificial}
Straw I, Callison-Burch C.
\newblock Artificial intelligence in mental health and the biases of language based models.
\newblock {\em PloS one\/} {\bf 15} (2020) e0240376.

\bibitem[{Pennington et~al.(2014)Pennington, Socher, and Manning}]{pennington-etal-2014-glove}
Pennington J, Socher R, Manning C.
\newblock {G}lo{V}e: Global vectors for word representation.
\newblock {\em Proceedings of the 2014 Conference on Empirical Methods in Natural Language Processing ({EMNLP})\/} (Doha, Qatar: Association for Computational Linguistics) (2014), 1532--1543.
\newblock \doi{10.3115/v1/D14-1162}.

\bibitem[{Mikolov et~al.(2013)Mikolov, Chen, Corrado, and Dean}]{mikolov2013efficient}
[Dataset] Mikolov T, Chen K, Corrado G, Dean J.
\newblock Efficient estimation of word representations in vector space (2013).

\bibitem[{Welbl et~al.(2021)Welbl, Glaese, Uesato, Dathathri, Mellor, Hendricks et~al.}]{welbl2021challenges}
[Dataset] Welbl J, Glaese A, Uesato J, Dathathri S, Mellor J, Hendricks LA, et~al.
\newblock Challenges in detoxifying language models (2021).

\bibitem[{Park and Rudzicz(2022)}]{park2022detoxifying}
[Dataset] Park YA, Rudzicz F.
\newblock Detoxifying language models with a toxic corpus (2022).

\bibitem[{Rodrigues(2020)}]{RODRIGUES2020100005}
Rodrigues R.
\newblock Legal and human rights issues of ai: Gaps, challenges and vulnerabilities.
\newblock {\em Journal of Responsible Technology\/} {\bf 4} (2020) 100005.
\newblock \doi{https://doi.org/10.1016/j.jrt.2020.100005}.

\bibitem[{Bender et~al.(2021)Bender, Gebru, McMillan-Major, and Shmitchell}]{Bender}
Bender EM, Gebru T, McMillan-Major A, Shmitchell S.
\newblock On the dangers of stochastic parrots: Can language models be too big?
\newblock {\em Proceedings of the 2021 ACM Conference on Fairness, Accountability, and Transparency\/} (New York, NY, USA: Association for Computing Machinery) (2021), FAccT '21, 610–623.
\newblock \doi{10.1145/3442188.3445922}.

\bibitem[{Steyerberg and Harrell(2016)}]{Steyerberg2016PredictionMN}
Steyerberg EW, Harrell FE.
\newblock Prediction models need appropriate internal, internal-external, and external validation.
\newblock {\em Journal of clinical epidemiology\/} {\bf 69} (2016) 245--7.

\bibitem[{Wainer and Cawley(2021)}]{WAINER2021115222}
Wainer J, Cawley G.
\newblock Nested cross-validation when selecting classifiers is overzealous for most practical applications.
\newblock {\em Expert Systems with Applications\/} {\bf 182} (2021) 115222.
\newblock \doi{https://doi.org/10.1016/j.eswa.2021.115222}.

\bibitem[{Patwardhan et~al.(2023)Patwardhan, Marrone, and Sansone}]{info14040242}
Patwardhan N, Marrone S, Sansone C.
\newblock Transformers in the real world: A survey on nlp applications.
\newblock {\em Information\/} {\bf 14} (2023).
\newblock \doi{10.3390/info14040242}.

\bibitem[{Lin(2004)}]{rouge}
Lin CY.
\newblock {ROUGE}: A package for automatic evaluation of summaries.
\newblock {\em Text Summarization Branches Out\/} (Barcelona, Spain: Association for Computational Linguistics) (2004), 74--81.

\bibitem[{Papineni et~al.(2002)Papineni, Roukos, Ward, and Zhu}]{papineni}
Papineni K, Roukos S, Ward T, Zhu WJ.
\newblock {B}leu: a method for automatic evaluation of machine translation.
\newblock {\em Proceedings of the 40th Annual Meeting of the Association for Computational Linguistics\/} (Philadelphia, Pennsylvania, USA: Association for Computational Linguistics) (2002), 311--318.
\newblock \doi{10.3115/1073083.1073135}.

\bibitem[{Vokinger et~al.(2021)Vokinger, Feuerriegel, and Kesselheim}]{Vokinger}
Vokinger K, Feuerriegel S, Kesselheim A.
\newblock Mitigating bias in machine learning for medicine.
\newblock {\em Communications Medicine\/} {\bf 1} (2021).
\newblock \doi{10.1038/s43856-021-00028-w}.

\bibitem[{Montesinos~L{\'o}pez et~al.(2022)Montesinos~L{\'o}pez, Montesinos~L{\'o}pez, and Crossa}]{montesinos2022overfitting}
Montesinos~L{\'o}pez OA, Montesinos~L{\'o}pez A, Crossa J.
\newblock Overfitting, model tuning, and evaluation of prediction performance.
\newblock {\em Multivariate statistical machine learning methods for genomic prediction\/} (Springer) (2022), 109--139.

\bibitem[{Norori et~al.(2021)Norori, Hu, Aellen, Faraci, and Tzovara}]{NORORI2021100347}
Norori N, Hu Q, Aellen FM, Faraci FD, Tzovara A.
\newblock Addressing bias in big data and ai for health care: A call for open science.
\newblock {\em Patterns\/} {\bf 2} (2021) 100347.
\newblock \doi{https://doi.org/10.1016/j.patter.2021.100347}.

\bibitem[{Mehrabi et~al.(2022)Mehrabi, Morstatter, Saxena, Lerman, and Galstyan}]{mehrabi2022survey}
[Dataset] Mehrabi N, Morstatter F, Saxena N, Lerman K, Galstyan A.
\newblock A survey on bias and fairness in machine learning (2022).

\bibitem[{Olteanu et~al.(2016)Olteanu, Castillo, Diaz, and Kiciman}]{Olteanu}
Olteanu A, Castillo C, Diaz F, Kiciman E.
\newblock Social data: Biases, methodological pitfalls, and ethical boundaries.
\newblock {\em SSRN Electronic Journal\/}  (2016).
\newblock \doi{10.2139/ssrn.2886526}.

\bibitem[{El~Naqa et~al.(2021)El~Naqa, Li, Fuhrman, Hu, Gorre, Chen et~al.}]{Naqa}
El~Naqa I, Li H, Fuhrman J, Hu Q, Gorre N, Chen W, et~al.
\newblock Lessons learned in transitioning to ai in the medical imaging of covid-19.
\newblock {\em Journal of Medical Imaging (Bellingham, Wash.)\/} {\bf 8} (2021) 010902--10902.
\newblock \doi{10.1117/1.JMI.8.S1.010902}.
\newblock T32 EB002103/EB/NIBIB NIH HHS/United States.

\bibitem[{Panch et~al.(2019)Panch, Mattie, and Atun}]{Panch}
Panch T, Mattie H, Atun R.
\newblock Artificial intelligence and algorithmic bias: Implications for health systems.
\newblock {\em Journal of Global Health\/} {\bf 9} (2019).
\newblock \doi{10.7189/jogh.09.020318}.

\bibitem[{McCradden et~al.(2020)McCradden, Joshi, Mazwi, and Anderson}]{McCradden}
McCradden M, Joshi S, Mazwi M, Anderson J.
\newblock Ethical limitations of algorithmic fairness solutions in health care machine learning.
\newblock {\em The Lancet Digital Health\/} {\bf 2} (2020) e221--e223.
\newblock \doi{10.1016/S2589-7500(20)30065-0}.

\bibitem[{Kim et~al.(2018)Kim, Wattenberg, Gilmer, Cai, Wexler, Viegas et~al.}]{kim2018interpretability}
[Dataset] Kim B, Wattenberg M, Gilmer J, Cai C, Wexler J, Viegas F, et~al.
\newblock Interpretability beyond feature attribution: Quantitative testing with concept activation vectors (tcav) (2018).

\bibitem[{Pymetrics(2020)}]{Pymetrics}
[Dataset] Pymetrics.
\newblock audit-ai.
\newblock \url{https://github.com/pymetrics/audit-ai} (2020).

\bibitem[{Speicher et~al.(2018)Speicher, Heidari, Grgic-Hlaca, Gummadi, Singla, Weller et~al.}]{Speicher}
Speicher T, Heidari H, Grgic-Hlaca N, Gummadi KP, Singla A, Weller A, et~al.
\newblock A unified approach to quantifying algorithmic unfairness: Measuring individual \&amp;group unfairness via inequality indices.
\newblock {\em Proceedings of the 24th ACM SIGKDD International Conference on Knowledge Discovery \&amp; Data Mining\/} (New York, NY, USA: Association for Computing Machinery) (2018), KDD '18, 2239–2248.
\newblock \doi{10.1145/3219819.3220046}.

\bibitem[{Zemel et~al.(2013)Zemel, Wu, Swersky, Pitassi, and Dwork}]{Zemel2013LearningFR}
Zemel RS, Wu LY, Swersky K, Pitassi T, Dwork C.
\newblock Learning fair representations.
\newblock {\em International Conference on Machine Learning\/} (2013).

\bibitem[{Mongan et~al.(2020)Mongan, Moy, and Kahn}]{PMID33937821}
Mongan J, Moy L, Kahn CEJ.
\newblock Checklist for artificial intelligence in medical imaging (claim): A guide for authors and reviewers.
\newblock {\em Radiology. Artificial intelligence\/} {\bf 2} (2020) e200029.
\newblock \doi{10.1148/ryai.2020200029}.

\bibitem[{Liu et~al.(2020{\natexlab{a}})Liu, Cruz~Rivera, Moher, Calvert, and Denniston}]{PMID32908283}
Liu X, Cruz~Rivera S, Moher D, Calvert MJ, Denniston AK.
\newblock Reporting guidelines for clinical trial reports for interventions involving artificial intelligence: the consort-ai extension.
\newblock {\em Nature medicine\/} {\bf 26} (2020{\natexlab{a}}) 1364--1374.
\newblock \doi{10.1038/s41591-020-1034-x}.

\bibitem[{Soysal et~al.(2018)Soysal, Wang, Jiang, Wu, Pakhomov, Liu et~al.}]{PMID29186491}
Soysal E, Wang J, Jiang M, Wu Y, Pakhomov S, Liu H, et~al.
\newblock {CLAMP} - a toolkit for efficiently building customized clinical natural language processing pipelines.
\newblock {\em Journal of the American Medical Informatics Association : JAMIA\/} {\bf 25} (2018) 331--336.
\newblock \doi{10.1093/jamia/ocx132}.

\bibitem[{Liu et~al.(2020{\natexlab{b}})Liu, Rivera, Moher, Calvert, and Denniston}]{Lium3164}
Liu X, Rivera SC, Moher D, Calvert MJ, Denniston AK.
\newblock Reporting guidelines for clinical trial reports for interventions involving artificial intelligence: the consort-ai extension.
\newblock {\em BMJ\/} {\bf 370} (2020{\natexlab{b}}).
\newblock \doi{10.1136/bmj.m3164}.

\bibitem[{Soysal et~al.(2017)Soysal, Wang, Jiang, Wu, Pakhomov, Liu et~al.}]{Soysal}
Soysal E, Wang J, Jiang M, Wu Y, Pakhomov S, Liu H, et~al.
\newblock {CLAMP – a toolkit for efficiently building customized clinical natural language processing pipelines}.
\newblock {\em Journal of the American Medical Informatics Association\/} {\bf 25} (2017) 331--336.
\newblock \doi{10.1093/jamia/ocx132}.

\bibitem[{Asan et~al.(2019)Asan, Bayrak, and Choudhury}]{Asan}
[Dataset] Asan O, Bayrak E, Choudhury A.
\newblock Artificial intelligence and human trust in healthcare: Focus on clinicians (preprint) (2019).
\newblock \doi{10.2196/preprints.15154}.

\bibitem[{Varkey(2020)}]{Varkey}
Varkey B.
\newblock Principles of clinical ethics and their application to practice.
\newblock {\em Medical Principles and Practice\/} {\bf 30} (2020).
\newblock \doi{10.1159/000509119}.

\bibitem[{Linardatos et~al.(2020)Linardatos, Papastefanopoulos, and Kotsiantis}]{Linardatos2020ExplainableAA}
Linardatos P, Papastefanopoulos V, Kotsiantis SB.
\newblock Explainable ai: A review of machine learning interpretability methods.
\newblock {\em Entropy\/} {\bf 23} (2020).

\bibitem[{Herm et~al.(2023)Herm, Heinrich, Wanner, and Janiesch}]{Herm_2023}
Herm LV, Heinrich K, Wanner J, Janiesch C.
\newblock Stop ordering machine learning algorithms by their explainability! a user-centered investigation of performance and explainability.
\newblock {\em International Journal of Information Management\/} {\bf 69} (2023) 102538.
\newblock \doi{10.1016/j.ijinfomgt.2022.102538}.

\bibitem[{Shin(2021)}]{SHIN2021102551}
Shin D.
\newblock The effects of explainability and causability on perception, trust, and acceptance: Implications for explainable ai.
\newblock {\em International Journal of Human-Computer Studies\/} {\bf 146} (2021) 102551.
\newblock \doi{https://doi.org/10.1016/j.ijhcs.2020.102551}.

\bibitem[{Carrillo et~al.(2021)Carrillo, Cantú, and Noriega}]{carrillo2021individual}
[Dataset] Carrillo A, Cantú LF, Noriega A.
\newblock Individual explanations in machine learning models: A survey for practitioners (2021).

\bibitem[{{C3.ai}(Accessed 2023)}]{c3ai_lime}
[Dataset] {C3ai}.
\newblock {LIME (Local Interpretable Model-Agnostic Explanations)}.
\newblock \url{https://c3.ai/glossary/data-science/lime-local-interpretable-model-agnostic-explanations/} (Accessed 2023).

\bibitem[{Lundberg and Lee(2017)}]{lundberg2017unified}
[Dataset] Lundberg S, Lee SI.
\newblock A unified approach to interpreting model predictions (2017).

\bibitem[{Ribeiro et~al.(2016)Ribeiro, Singh, and Guestrin}]{Tulio}
[Dataset] Ribeiro MT, Singh S, Guestrin C.
\newblock "why should i trust you?": Explaining the predictions of any classifier (2016).

\bibitem[{Khanmohammadi et~al.(2021{\natexlab{b}})Khanmohammadi, Mirshafiee, Ghassemi, and Alhanai}]{fetal}
[Dataset] Khanmohammadi R, Mirshafiee MS, Ghassemi MM, Alhanai T.
\newblock Fetal gender identification using machine and deep learning algorithms on phonocardiogram signals (2021{\natexlab{b}}).

\bibitem[{Goodfellow et~al.(2018)Goodfellow, McDaniel, and Papernot}]{Goodfellow}
Goodfellow I, McDaniel P, Papernot N.
\newblock Making machine learning robust against adversarial inputs.
\newblock {\em Communications of the ACM\/} {\bf 61} (2018) 56--66.
\newblock \doi{10.1145/3134599}.

\bibitem[{Biggio et~al.(2013)Biggio, Corona, Maiorca, Nelson, {\v{S} }rndi{\'{c}}, Laskov et~al.}]{Biggio_2013}
Biggio B, Corona I, Maiorca D, Nelson B, {\v{S} }rndi{\'{c}} N, Laskov P, et~al.
\newblock Evasion attacks against machine learning at test time.
\newblock {\em Advanced Information Systems Engineering\/} (Springer Berlin Heidelberg) (2013), 387--402.
\newblock \doi{10.1007/978-3-642-40994-3_25}.

\bibitem[{Borkar and Chen(2021)}]{borkar2021simple}
[Dataset] Borkar J, Chen PY.
\newblock Simple transparent adversarial examples (2021).

\bibitem[{Drabiak et~al.(2023)Drabiak, Kyzer, Nemov, and El~Naqa}]{Drabiak}
Drabiak K, Kyzer S, Nemov V, El~Naqa I.
\newblock Ai and ml ethics, law, diversity, and global impact.
\newblock {\em The British journal of radiology\/}  (2023) 20220934.
\newblock \doi{10.1259/bjr.20220934}.

\bibitem[{{U.S. Food and Drug Administration}(2021{\natexlab{a}})}]{FDAAIML}
[Dataset] {US Food and Drug Administration}.
\newblock Artificial intelligence and machine learning discussion paper.
\newblock [Online] (2021{\natexlab{a}}).
\newblock Accessed: May 30, 2023.

\bibitem[{{U.S. Food and Drug Administration}(2021{\natexlab{b}})}]{FDA2023}
[Dataset] {US Food and Drug Administration}.
\newblock {Artificial Intelligence and Machine Learning Software as a Medical Device}.
\newblock \url{https://www.fda.gov/medical-devices/software-medical-device-samd/artificial-intelligence-and-machine-learning-software-medical-device} (2021{\natexlab{b}}).
\newblock Accessed: May 30, 2023.

\bibitem[{{U.S. Food and Drug Administration}(2021{\natexlab{c}})}]{FDA2023b}
[Dataset] {US Food and Drug Administration}.
\newblock {Good Machine Learning Practice for Medical Device Development: Guiding Principles}.
\newblock \url{https://www.fda.gov/medical-devices/software-medical-device-samd/good-machine-learning-practice-medical-device-development-guiding-principles} (2021{\natexlab{c}}).
\newblock Accessed: May 30, 2023.

\bibitem[{Gilbert et~al.(2023)Gilbert, Harvey, Melvin, Vollebregt, and Wicks}]{Gilbert2023}
Gilbert S, Harvey H, Melvin T, Vollebregt E, Wicks P.
\newblock Large language model ai chatbots require approval as medical devices.
\newblock {\em Nature Medicine\/} {\bf 29} (2023) 2396--2398.
\newblock \doi{10.1038/s41591-023-02412-6}.

\bibitem[{Naik et~al.(2022)Naik, Hameed, Shetty, Swain, Shah, Paul et~al.}]{Naik}
Naik N, Hameed B, Shetty D, Swain D, Shah M, Paul R, et~al.
\newblock Legal and ethical consideration in artificial intelligence in healthcare: Who takes responsibility?
\newblock {\em Frontiers in Surgery\/} {\bf 9} (2022) 862322.
\newblock \doi{10.3389/fsurg.2022.862322}.

\bibitem[{Smith(2021)}]{Helen}
Smith H.
\newblock Clinical ai: opacity, accountability, responsibility and liability.
\newblock {\em AI \& SOCIETY\/} {\bf 36} (2021).
\newblock \doi{10.1007/s00146-020-01019-6}.

\bibitem[{Smith and Fotheringham(2020)}]{Fotheringham}
Smith H, Fotheringham K.
\newblock Artificial intelligence in clinical decision-making: Rethinking liability.
\newblock {\em Medical Law International\/} {\bf 20} (2020) 131--154.
\newblock \doi{10.1177/0968533220945766}.

\bibitem[{Gerke et~al.(2020)Gerke, Minssen, and Cohen}]{Gerke2020}
Gerke S, Minssen T, Cohen G.
\newblock {\em Ethical and legal challenges of artificial intelligence-driven healthcare\/} (United States: Elsevier Inc.) (2020), 295--336.
\newblock \doi{10.1016/B978-0-12-818438-7.00012-5}.

\bibitem[{Amurao et~al.(2023)Amurao, Gress, Keenan, Halvorsen, Nye, and Mahesh}]{Amurao}
Amurao M, Gress D, Keenan M, Halvorsen P, Nye J, Mahesh M.
\newblock Quality management, quality assurance, and quality control in medical physics.
\newblock {\em Journal of Applied Clinical Medical Physics\/} {\bf 24} (2023).
\newblock \doi{10.1002/acm2.13885}.

\bibitem[{Vandewinckele et~al.(2020)Vandewinckele, Claessens, Dinkla, Brouwer, Crijns, Verellen et~al.}]{VANDEWINCKELE202055}
Vandewinckele L, Claessens M, Dinkla A, Brouwer C, Crijns W, Verellen D, et~al.
\newblock Overview of artificial intelligence-based applications in radiotherapy: Recommendations for implementation and quality assurance.
\newblock {\em Radiotherapy and Oncology\/} {\bf 153} (2020) 55--66.
\newblock \doi{https://doi.org/10.1016/j.radonc.2020.09.008}.
\newblock Physics Special Issue: ESTRO Physics Research Workshops on Science in Development.

\end{thebibliography}
\end{document}